\documentclass[10pt, a4paper]{article}
\usepackage{lrec}
\usepackage{graphicx}
\usepackage{tabularx}
\usepackage{soul}

\usepackage{titlesec}
\titleformat{\section}{\normalfont\large\bf\center}{\thesection.}{1em}{}
\titleformat{\subsection}{\normalfont\SmallTitleFont\bf\raggedright}{\thesubsection.}{1em}{}
\titleformat{\subsubsection}{\normalfont\normalsize\bf\raggedright}{\thesubsubsection.}{1em}{}
\renewcommand\thesection{\arabic{section}}
\renewcommand\thesubsection{\thesection.\arabic{subsection}}
\renewcommand\thesubsubsection{\thesubsection.\arabic{subsubsection}}

\usepackage{epstopdf}
\usepackage[latin1]{inputenc}

\usepackage{hyperref}
\usepackage{xstring}
\usepackage{comment}
\usepackage{xspace}
\usepackage{caption}
\usepackage{tikz}
\usepackage{graphicx}
\usepackage{pgfplots}
\usepackage{setspace}

\usepackage{hyperref}
\usepackage{xstring}

\usepackage{url}
\usepackage{subcaption}

\title{A Short Survey on Sense-Annotated Corpora}

\name{Tommaso Pasini$^{1}$, Jose Camacho-Collados$^{2}$}

\address{$^1$Department of Computer Science, Sapienza University of Rome, Italy \\ $^2$School of Computer Science and Informatics, Cardiff University, United Kingdom \\
         \tt $^1$pasini@di.uniroma1.it, \tt $^2$camachocolladosj@cardiff.ac.uk 
         }

\abstract{
Large sense-annotated datasets are increasingly necessary for training deep supervised systems in Word Sense Disambiguation. However, gathering high-quality sense-annotated data for as many instances as possible is a laborious and expensive task. This has led to the proliferation of automatic and semi-automatic methods for overcoming the so-called knowledge-acquisition bottleneck. In this short survey we present an overview of sense-annotated corpora, annotated either manually- or (semi)automatically, that are currently available for different languages and featuring distinct lexical resources as inventory of senses, i.e. WordNet, Wikipedia, BabelNet. Furthermore, we provide the reader with general statistics of each dataset and an analysis of their specific features.  \\ \newline  }
\pgfplotsset{compat=1.14}

\begin{document}

\maketitleabstract





\section{Introduction}
Word Sense Disambiguation (WSD) is a key task in Natural Language Understanding. It consists in assigning the appropriate meaning from a pre-defined sense inventory to a word in context\cite{navigli:09}. While knowledge-based approaches to this task have been proposed \cite{agirreetal:14,Moroetal:14tacl,butnaru-ionescu-hristea:2017:EACLlong,chaplot2018knowledge}, supervised approaches \cite{zhongng:10,melamudetal:16,iacobaccietal:16,kagelback:2016,luo2018incorporating} have been more effective in terms of performance when sense-annotated corpora are available \cite{raganatoetal:17,huang2019glossbert}. Unfortunately, obtaining such data is heavily time-consuming and expensive \cite{schubert:2006}, and a reasonable amount of manually-annotated instances are available for English only \cite{Milleretal:93}. 

One of the first attempts towards building large sense-annotated corpora was SemCor \cite{Milleretal:93} which contains instances annotated with senses from WordNet \cite{Fellbaum:98}. Since then, several semi-automatic and automatic approaches have also been proposed \cite{taghipourng:15,dellibovietal:17,pasiniandnavigli:18}. These automatic efforts tend to produce noisier annotations, but their coverage has been shown to lead to better supervised and semi-supervised WSD systems \cite{taghipour2015semi,otegi:2016,raganatoetal:16,yuan:16,dellibovietal:17,pasininavigli:17}, as well as to learn effective embedded representations for senses \cite{iacobacci:2015,flekova:2016,mancini2017sw2v}. Nevertheless, each of the aforementioned datasets come with its own format, hence making it complicated to merge them or moving from one to another. \newcite{vialetal:18} tackled this specific problem and, following \newcite{raganatoetal:17}, proposed UFSAC, a unified repository of several sense-annotated corpora all with the same format, thus making it easier develop and test WSD models on different datasets.

In this survey we present the main approaches in the literature to build sense-annotated corpora, not only for WordNet but also for multilingual sense inventories, namely Wikipedia and BabelNet. There have been additional constructing sense-annotated data for other resources such as the New Oxford American Dictionary \cite{yuan:16} or other language-specific versions like GermaNet \cite{henrich2012webcage}. While these language-specific resources are certainly relevant, in this paper we have focused on English WordNet and multilingual resources with a higher coverage like Wikipedia and BabelNet.\footnote{For a more specific survey on corpora annotated with language-specific versions of WordNet, please refer to \newcite{petrolito2014survey}.} Finally, we provide a general overview and statistics of these sense-annotated resources, providing relevant details 
across resources and languages.

\section{Sense-Annotated Corpora}

\begin{figure*}[t!]

\begin{center}
		\includegraphics[scale=0.325]{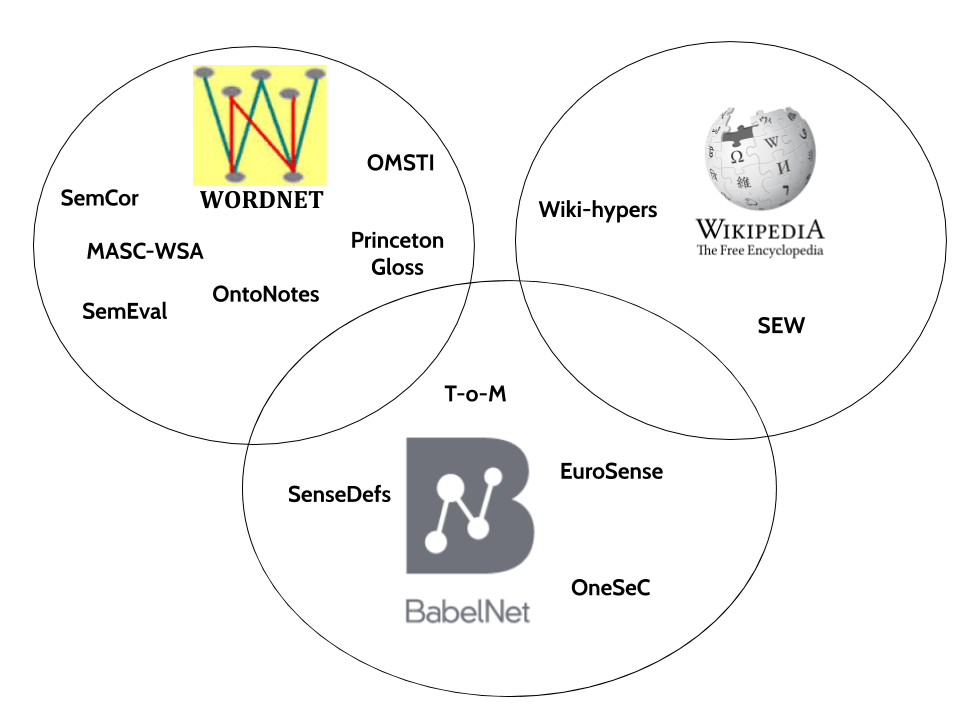}
\end{center}
    \caption{Overview of sense inventories with their corresponding sense-annotated corpora.}
    \label{fig:flow}
\end{figure*}

In this Section we describe the main efforts compiling sense-annotated corpora. 
We present currently available corpora for three resources: WordNet (Section \ref{wordnet}), Wikipedia (Section \ref{wikipedia}) and BabelNet (Section \ref{babelnet}). Figure \ref{fig:flow} provides an overview of these resources and their underlying corpora. 

\subsection{WordNet}
\label{wordnet}

WordNet \cite{Fellbaum:98} has been one of the most widely used knowledge resource in lexical semantics. In fact, it is the \textit{de-facto} standard sense inventory for Word Sense Disambiguation since many years. The core unit in WordNet is the synset. A synset represents a concept or a meaning which is represented by its various lexicalizations (i.e. senses). For example, the synset defined as \textit{motor vehicle with four wheels} can be expressed by its synonym senses \textit{auto, automobile, machine} and  \textit{motorcar}. In what follows we list the main WordNet sense-annotated corpora, using WordNet 3.0 as reference sense inventory.

\paragraph{SemCor.} The first and most prominent example of sense-annotated corpora is SemCor \cite{miller93}. SemCor was manually annotated and consists of 352 documents from the Brown Corpus \cite{kucera1979standard} and 226,040 sense annotations. SemCor is the largest manually-annotated corpus and the most used in the literature to train WSD supervised systems \cite{agirre2009semeval,zhongng:10,raganatoemnlp2017,luo2018incorporating,loureiro-jorge-2019-language,huang2019glossbert}.

\paragraph{SemEval.} SemEval datasets provide reliable benchmarks for testing WSD systems. The main datasets from Senseval and SemEval competitions have been compiled and unified by \newcite{raganatoetal:17}. In particular, the datasets from Senseval-2 \cite{EdmondsCotton:01}, Senseval-3 task 1 \cite{SnyderPalmer:04}, SemEval-2007 task 17 \cite{Pradhanetal:07}, SemEval-2013 task 12 \cite{Naviglietal:13}, and SemEval-2015 task 13 \cite{MoroNavigli:15}. These datasets, which have been mainly used as evaluation benchmarks for WSD systems, contain a total of 7,253 sense annotations.

\paragraph{MASC-WSA.} The MASC Word Sense Annotation (MASC-WSA) corpus \cite{ide-etal-2010-manually}
is an excerpt of the Manually Annotated Sub-Corpus of American English \cite[MASC]{ideetal:08} and the Open American National Corpus \cite[ANC]{ideetal:02} containing annotations for 45 distinct lexemes, i.e., lemma-pos pairs, for a total of 441 distinct WordNet word senses.\footnote{This corpus has also been annotated with other resources, such as FrameNet \cite{Bakeretal:98}. A comparison between the sense annotations of WordNet and lexical units of FrameNet is provided in \newcite{demeloetal:12}.} Each word occurrence has been manually annotated on Amazon Mechanichal Turk by roughly 25 persons for a total of 1M annotations.

\paragraph{Princeton WordNet Gloss.}
The Princeton WordNet Gloss Corpus\footnote{\url{http://wordnet.princeton.edu/glosstag.shtml}} is a sense-annotated corpus of textual definitions (glosses) from WordNet synsets. The corpus was tagged semi-automatically: 330,499 instances were annotated manually while the remaining annotations (i.e.~118,856) were obtained automatically. This corpus of disambiguated glosses has already proved to be useful in tasks such as semantic similarity \cite{pilehvaretal:13}, domain labeling \cite{gonzalez2012graph} and Word Sense Disambiguation \cite{baldwin2008mrd,agirre09,camachocolladosetal:2015b}.




\paragraph{OntoNotes.}
OntoNotes \cite{weischedel2013ontonotes} is a corpus from the Linguistic Data Consortium which comprises different kinds of explicitly-tagged syntactic and semantic information, including annotations at the sense level. The OntoNotes corpus consists of documents from diverse genres such as news, weblogs and telephone conversation. Its 5.0 released version contains 264,622 sense annotations.  

\paragraph{OMSTI.} The task of gathering sense annotations has proved expensive and not easily scalable. That is the reason why more recent approaches have attempted to exploit semi-automatic or automatic techniques. OMSTI\footnote{\url{http://www.comp.nus.edu.sg/~nlp/corpora.html}} 
\cite[One Million Sense-Tagged Instances]{taghipourng:15}, which is a semi-automatically constructed corpus annotated with WordNet senses, is a prominent example. It was built by exploiting the alignment-based WSD approach of \newcite{chan2005scaling} on a large English-Chinese parallel corpus \cite[MultiUN corpus]{multiUN2010}. OMSTI
, coupled with SemCor, has already been successfully leveraged as training data for training supervised systems \cite{taghipourng:15,iacobaccietal:16,raganatoetal:17}.

\subsection{Wikipedia}
\label{wikipedia}

Wikipedia is a collaboratively-constructed encyclopedic resource representing concepts and entities with a so-called Wikipedia article. In addition to a large coverage of concepts and entities, Wikipedia provides multilinguality, as it covers over 250 languages and these languages are connected via interlingual links. In this Section we describe two datasets providing disambiguations in the form of Wikipedia pages.\footnote{Note that more Wikipedia sense-annotated datasets extracted from the Wikilinks project exist \cite{singh2012wikilinks,noisynedconll2017}. However, due to privacy and license issues, these datasets cannot be shared directly. Please also refer to \newcite{usbeck2015gerbil} for an overview and unification of datasets focused on Entity Linking. 
} For these two datasets we have used the same version of Wikipedia for a more accurate comparison\footnote{We considered the Wikipedia dumps of November 2014.}.

\paragraph{Wikipedia hyperlinks.} This corpus contains the full Wikipedia multilingual corpus with hyperlinks as sense-annotated instances. Hyperlinks are highlighted mentions within a Wikipedia article that directly links to another Wikipedia page. 

\paragraph{SEW.}
The Semantically Enriched Wikipedia \footnote{\url{http://lcl.uniroma1.it/sew}}
\cite[SEW]{raganatoetal:16} is a Wikipedia-sense annotated corpus which was constructed by exploiting Wikipedia hyperlinks, propagating them across Wikipedia pages.  
Its English version comprises over 160M sense annotations with an estimated precision over 90\%.

\begin{table*}

\begin{center}
{

\renewcommand{\arraystretch}{1.2}
\resizebox{\linewidth}{!}{ 
\begin{tabular}{|l|c|c|r|r||r|r|r|r|}

\cline{6-9}

 \multicolumn{1}{l}{}  &  \multicolumn{1}{c}{} & \multicolumn{1}{c}{} &  \multicolumn{1}{c}{} & \multicolumn{1}{c}{} & \multicolumn{4}{|c|}{\textbf{English}} \\

\cline{2-9}

\multicolumn{1}{l|}{}  &\multicolumn{1}{c|}{\textbf{Resource}} &
\multicolumn{1}{c|}{\textbf{Type}} & \multicolumn{1}{c|}{\textbf{\#Langs}}	&  
\multicolumn{1}{c||}{\bf \#Annotations}  &\multicolumn{1}{c|}{\bf \#Tokens}  &\multicolumn{1}{c|}{\bf \#Annotations}  &\multicolumn{1}{c|}{\bf Amb} & \multicolumn{1}{c|}{\bf Entropy}    \\
\hline


\textbf{SemCor} & WordNet & M &  1  
&  226,036  &  802,443  	&  226,036  & 6.8    & 0.27   \\
\hline
\textbf{MASC-WSA} & WordNet & M & 1 & 1,084,551 & 1,309,838 & 1,084,551 & 8.8 & 2.09\\
\hline
\textbf{SemEval-ALL} & WordNet & M &  1  
&  7,253  &  25,503  	&  7,253 & 5.8    & 0.18 \\
\hline
\textbf{OntoNotes} & WordNet & M &  1  
&  264,622  &  1,445,000 	&  264,622  & - &-\\ 
\hline
\textbf{Princeton Gloss} & WordNet & SA &  1  
&  449,355  &  1,621,129 	&  449,355  & 3.8 & 0.45\\ 
\hline
\textbf{OMSTI} & WordNet & SA & 1 
&  911,134  &  30,441,386 	&  911,134  & 8.9   & 0.94 \\


\hline
\hline

\textbf{Wiki-hypers} & Wikipedia & C &  271  
&  321,718,966  &  1,357,105,761  	&  71,457,658 
& 2.6 & 0.44 \\
\hline
\textbf{SEW} & Wikipedia & SA &  1  
&  162,614,753  &  1,357,105,761 	& 162,614,753  & 7.9 & 0.40\\ 

\hline
\hline

\textbf{SenseDefs} & BabelNet & A &  263  
& 163,029,131  &  71,109,002 	&  37,941,345  & 4.6    &0.04\\ 
\hline
\textbf{EuroSense} & BabelNet & A &  21 
& 122,963,111  &  48,274,313	&  15,502,847  & 6.5   & 0.21\\
\hline
\textbf{T-o-M} & BabelNet & A &  6  
&  17,987,488  & 291,550,966 &  12,722,530  & 3.6  & 0.48  \\
\hline
\textbf{OneSeC} & BabelNet & A & 5 & 1,222,090 & 25,017,839 & 888,417 & 3.6 & 0.51 \\ 
\hline
\end{tabular}
}
}
\end{center}
\caption{\label{tab:stats} Statistics of the sense-annotated corpora across languages and resources. Type ``M" stands for Manual, ``SA" stands Semi-automatic, ``C" for Collaborative and ``A" for Automatic.}
\end{table*}

\subsection{BabelNet}
\label{babelnet}

BabelNet \cite{NavigliPonzetto:12aij} is a wide-coverage multilingual semantic network obtained from the  integration of various encyclopedias and dictionaries (WordNet and Wikipedia, inter alia).
Being a superset of all these resources, BabelNet brings together lexicographic and encyclopedic knowledge, thus containing both named entities and concepts, and, unlike Wikipedia covering only noun instances, instances have diverse Part-Of-Speech (PoS) tags: nouns, verbs, adjectives and adverbs. Given its multilingual nature (i.e. BabelNet covers over 250 languages), BabelNet has been used as a sense inventory for annotating text in languages other than English. 

\paragraph{SenseDefs.} SenseDefs\footnote{\url{http://lcl.uniroma1.it/sensedefs}}
\cite{camacho2019sensedefs} extends the effort from the Princeton WordNet Gloss Corpus project (see Section \ref{wordnet}) by automatically disambiguating textual definitions from various heterogeneous sources in 263 languages. The underlying idea lies on leveraging the cross-complementarities of definitions of identical concepts from different languages and resources. The approach couples a graph-based disambiguation method \cite{Moroetal:14tacl} with a refinement based on distributional similarity \cite{camacho2016nasari}. The proposed method was evaluated on four European languages (English, Spanish, French and Italian) with an estimated precision of over 80\%.

\paragraph{EuroSense.} The construction of EuroSense\footnote{\url{http://lcl.uniroma1.it/eurosense}} 
\cite{dellibovietal:17} follows a similar approach to SenseDefs. In this case, parallel corpora is exploited for a single multilingual disambiguation. The output is a sense-annotated corpus for 21 languages for the Europarl parallel corpus \cite{koehn05}. The estimated precision for four languages isover 80\% on average, with a peak of almost 90\% for German.

\paragraph{Train-o-Matic.} Similarly to the previous approach, Train-o-Matic\footnote{\url{http://trainomatic.org}} 
\cite[T-o-M]{pasininavigli:17} aims at automatically annotating words from a raw corpus with senses. The main difference with respect to EuroSense and OMSTI lies in the fact that T-o-M  does not need parallel data in order to annotate the input corpus. While being language independent and fully automatic, it proved to lead supervised systems to high performance, close or even better than those achieved when a manually annotated corpus (e.g. SemCor) is used for training. Moreover, it has also proved effective in languages other than English \cite{pasinietal:18}: Italian, Spanish, French, German and Chinese. 
\paragraph{OneSeC.}
OneSeC\footnote{\url{http://trainomatic.org/onesec}} \cite{scarlinietal:19} is the most recent work among those aiming at automatically producing semantically-annotated data. Instead of the well-known ``one sense per discourse`` assumption made by
\newcite{galeetal:92}, this work makes a more relaxed hypotesis, i.e., ``one sense per Wikipedia Category``. That is, a noun is used always with the same meaning within a Wikipedia Category. By leveraging this conjecture, OneSeC exploits the texts contained within the pages of a given Wikipedia Category to annotate each noun occurrence therein with its most suitable meaning. The corpora for English showed to be of high-quality, leading a supervised English WSD model, i.e., It Makes Sense \cite[IMS]{zhongng:10}, to achieve results that are higher than those attained by IMS trained on other automatically generated corpora. Furthermore, OneSeC has been used to generate annotated data for four other European languages, namely: Italian, Spanish, German and French. 
\section{Statistics}

In order to have a global overview of all sense-annotated corpora, their main features are displayed in Table \ref{tab:stats}. For each corpus we include its underlying resource, number of languages covered and total number of sense annotations. In general the datasets are quite heterogeneous in nature, coming from three different resources and constructed via four different strategies: manual, semi-automatic, automatic and collaborative. The number of sense annotations also varies depending on the resource, with Wikipedia- and BabelNet-based corpora contributing with the highest number of annotations. This is correlated with the coverage of these resources: Wikipedia and BabelNet are two orders of magnitude higher than WordNet.

In addition to these global statistics, Table \ref{tab:stats} shows local statistics (i.e. number of tokens, number of sense annotations, ambiguity level and entropy) for English, which is the only language covered by all corpora.\footnote{Due to license restrictions we could not access OntoNotes' full corpus for computing its ambiguity/entropy.} The ambiguity level of each dataset is computed as the average number of candidate senses per instance (i.e., senses with the same surface form of a target word). 
The average entropy score is first computed separately for each word sense distribution of frequencies, and then averaged by the number of unique lemmas in the dataset \cite{pasiniandnavigli:18}. These two measures are complementary: 
while the average ambiguity is a precise indicator of the number of senses associated to a word within the corpus, the average entropy represent the skewness of word senses distributions, with higher entropy scores meaning that the sense distributions are closer to the uniform distribution.

\section{Analysis} 

\pgfplotsset{
    yticklabel style={/pgf/number format/fixed},  
}

\begin{figure}[!t]
    \begin{tikzpicture}

        \begin{axis}[
                legend pos=north west,
            xtick=data,
            xticklabels={Wiki-hy,T-o-M, OneSeC, P. Gloss, SenseDefs, SemEval-ALL, EuroSense, SemCor, SEW, MASC-WSA, OMSTI},
            xticklabel style = {rotate=45,anchor=east},
            ]
            \addplot[orange,mark=square*] table [y=Polisemy]{thu2.dat};
            \addlegendentry{Polisemy}
            \addplot[green,mark=square*] table [y= Entropy]{thu2.dat};
            \addlegendentry{Entropy}
            \end{axis}
        \end{tikzpicture}
            \caption{Normalized entropy and ambiguity levels across datasets.}
            \label{fig:polvsent}

\end{figure}
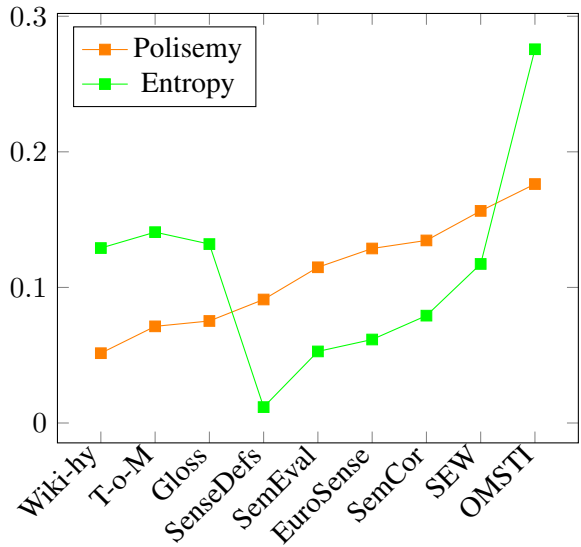
%

To gain insights on the features of each sense-annotated corpora, we provide a small analysis on the entropy and ambiguity levels. Figure \ref{fig:polvsent} shows the average ambiguity of a dataset (x axis) with respect to the average entropy (y axis) of the lemma annotations therein\footnote{Average ambiguity and entropy values shown in Table \ref{tab:stats} are normalized by the sum of all ambiguity or entropy values.}. As can be observed, datasets with higher degree of ambiguity tend to be also more entropic with MASC-WSA being the corpus with the least skewed sense distributions.\footnote{The high entropy of MASC-WSA may also be attributed to the specific selection of the vocabulary, which was decided by a committee from WordNet and FrameNet management teams.}
On the other hand, datasets with lower levels of ambiguity tend to have more unbalanced distributions and consequently a lower degree of entropy. For instance, EuroSense, which was automatically-constructed, have the most similar entropy to that of SemCor and SemEval datasets, which were manually-curated. 
We note that SenseDefs is the dataset with the lowest entropy. Going more in-depth we observed that, due to its nature, SenseDefs contains a large number of unambiguous named entities, i.e., containing a single sense in its underlying sense inventory BabelNet. 

\section{Conclusion}
In this paper we have presented an overview of available sense-annotated datasets for WordNet, Wikipedia and BabelNet, and for various languages. These datasets correspond to a wide variety of approaches, from manual construction to automatic or semi-automatic methods. By listing and providing statistics for all these datasets we are pursuing two main goals: (1) motivating and providing information about sense-annotated corpora to be used for research purposes, and (2) highlighting the main properties of the various sense-annotated corpora across resources. 

Moreover, this paper represents a first step for obtaining a fully-integrated repository of sense-annotated corpora which can be easily leveraged for research and evaluation purposes. As future work, we plan to integrate these sense-annotated resources into a unified multilingual repository, following the lines of \newcite{raganatoetal:17} and \newcite{vialetal:18} for WordNet sense-annotated corpora in English. 


\section*{Acknowledgments}

The authors would like to thank Claudio Delli Bovi and Alessandro Raganato for interesting discussions and their input with some of the datasets.

\noindent
\begin{figure}[!ht]
\begin{subfigure}{0.1\columnwidth}
    \includegraphics[scale=0.8]{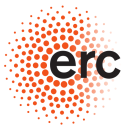}
\end{subfigure}
\hspace{0.01\linewidth}
\begin{subfigure}{0.72\columnwidth}
  Tommaso Pasini gratefully acknowledges the support of the ERC Consolidator Grant MOUSSE No. 726487 under the European
Union's Horizon 2020 research and innovation programme.
\end{subfigure}
\hspace{0.01\linewidth}
\begin{subfigure}{0.05\columnwidth}
\includegraphics[scale=0.25]{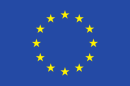}
\end{subfigure}
\end{figure}

\section{Bibliographical References}
\label{main:ref}

\bibliographystyle{lrec}
\bibliography{lrec2020W-xample}

\begin{thebibliography}{}

\bibitem[\protect\citename{Agirre and Soroa}2009]{agirre09}
Agirre, E. and Soroa, A.
\newblock (2009).
\newblock Personalizing {P}age{R}ank for {W}ord {S}ense {D}isambiguation.
\newblock In {\em Proceedings of the 12th Conference of the European Chapter of
  the Association for Computational Linguistics, {Athens, Greece, 30 March--3
  April 2009}}, pages 33--41.

\bibitem[\protect\citename{Agirre \bgroup et al.\egroup
  }2009]{agirre2009semeval}
Agirre, E., De~Lacalle, O.~L., Fellbaum, C., Marchetti, A., Toral, A., and
  Vossen, P.
\newblock (2009).
\newblock Semeval-2010 task 17: All-words word sense disambiguation on a
  specific domain.
\newblock In {\em Proceedings of the Workshop on Semantic Evaluations: Recent
  Achievements and Future Directions}, pages 123--128. Association for
  Computational Linguistics.

\bibitem[\protect\citename{Agirre \bgroup et al.\egroup }2014]{agirreetal:14}
Agirre, E., de~Lacalle, O.~L., and Soroa, A.
\newblock (2014).
\newblock Random walks for knowledge-based word sense disambiguation.
\newblock {\em Computational Linguistics}, 40(1):57--84.

\bibitem[\protect\citename{Baker \bgroup et al.\egroup }1998]{Bakeretal:98}
Baker, C.~F., Fillmore, C.~J., and Lowe, J.~B.
\newblock (1998).
\newblock {The Berkeley FrameNet project}.
\newblock In {\em Proceedings of the 36th Annual Meeting of the Association for
  Computational Linguistics and 17th International Conference on Computational
  Linguistics (COLING-ACL '98)}, pages 86--90, Montreal, Canada.

\bibitem[\protect\citename{Baldwin \bgroup et al.\egroup }2008]{baldwin2008mrd}
Baldwin, T., Su, N.~K., Bond, F., Fujita, S., Martinez, D., and Tanaka, T.
\newblock (2008).
\newblock Mrd-based word sense disambiguation: Further extending lesk.
\newblock In {\em Proceedings of International Joint Conference on Natural
  Language Processing}, pages 775--780.

\bibitem[\protect\citename{Butnaru \bgroup et al.\egroup
  }2017]{butnaru-ionescu-hristea:2017:EACLlong}
Butnaru, A., Ionescu, R.~T., and Hristea, F.
\newblock (2017).
\newblock Shotgunwsd: An unsupervised algorithm for global word sense
  disambiguation inspired by dna sequencing.
\newblock In {\em Proceedings of the 15th Conference of the European Chapter of
  the Association for Computational Linguistics: Volume 1, Long Papers}, pages
  916--926, Valencia, Spain. Association for Computational Linguistics.

\bibitem[\protect\citename{Camacho-Collados \bgroup et al.\egroup
  }2015]{camachocolladosetal:2015b}
Camacho-Collados, J., Pilehvar, M.~T., and Navigli, R.
\newblock (2015).
\newblock A unified multilingual semantic representation of concepts.
\newblock In {\em Proceedings of ACL}, pages 741--751, Beijing, China, July.

\bibitem[\protect\citename{Camacho-Collados \bgroup et al.\egroup
  }2016]{camacho2016nasari}
Camacho-Collados, J., Pilehvar, M.~T., and Navigli, R.
\newblock (2016).
\newblock {Nasari: Integrating explicit knowledge and corpus statistics for a
  multilingual representation of concepts and entities}.
\newblock {\em Artificial Intelligence}, 240:36--64.

\bibitem[\protect\citename{Camacho-Collados \bgroup et al.\egroup
  }2019]{camacho2019sensedefs}
Camacho-Collados, J., Bovi, C.~D., Raganato, A., and Navigli, R.
\newblock (2019).
\newblock Sensedefs: a multilingual corpus of semantically annotated textual
  definitions.
\newblock {\em Language Resources and Evaluation}, 53(2):251--278.

\bibitem[\protect\citename{Chan and Ng}2005]{chan2005scaling}
Chan, Y.~S. and Ng, H.~T.
\newblock (2005).
\newblock Scaling up word sense disambiguation via parallel texts.
\newblock In {\em AAAI}, volume~5, pages 1037--1042.

\bibitem[\protect\citename{Chaplot and
  Salakhutdinov}2018]{chaplot2018knowledge}
Chaplot, D.~S. and Salakhutdinov, R.
\newblock (2018).
\newblock Knowledge-based word sense disambiguation using topic models.
\newblock In {\em Proceedings of AAAI}.

\bibitem[\protect\citename{De~Melo \bgroup et al.\egroup }2012]{demeloetal:12}
De~Melo, G., Baker, C.~F., Ide, N., Passonneau, R.~J., and Fellbaum, C.
\newblock (2012).
\newblock Empirical comparisons of masc word sense annotations.
\newblock In {\em LREC}, pages 3036--3043.

\bibitem[\protect\citename{Delli~Bovi \bgroup et al.\egroup
  }2017]{dellibovietal:17}
Delli~Bovi, C., Camacho-Collados, J., Raganato, A., and Navigli, R.
\newblock (2017).
\newblock {EuroSense}: Automatic harvesting of multilingual sense annotations
  from parallel text.
\newblock In {\em Proc.of ACL}, volume~2, pages 594--600.

\bibitem[\protect\citename{Edmonds and Cotton}2001]{EdmondsCotton:01}
Edmonds, P. and Cotton, S.
\newblock (2001).
\newblock Senseval-2: overview.
\newblock In {\em Proc. of SensEval 2}, pages 1--5. ACL.

\bibitem[\protect\citename{Eisele and Chen}2010]{multiUN2010}
Eisele, A. and Chen, Y.
\newblock (2010).
\newblock {MultiUN: A Multilingual Corpus from United Nation Documents}.
\newblock In {\em Proceedings of the Seventh conference on International
  Language Resources and Evaluation}, pages 2868--2872.

\bibitem[\protect\citename{Eshel \bgroup et al.\egroup
  }2017]{noisynedconll2017}
Eshel, Y., Cohen, N., Radinsky, K., Markovitch, S., Yamada, I., and Levy, O.
\newblock (2017).
\newblock Named entity disambiguation for noisy text.
\newblock In {\em Proceedings of the 21st Conference on Computational Natural
  Language Learning (CoNLL 2017)}, pages 58--68. Association for Computational
  Linguistics.

\bibitem[\protect\citename{Fellbaum}1998]{Fellbaum:98}
Fellbaum, C.
\newblock (1998).
\newblock {\em {W}ord{N}et: An Electronic Database}.
\newblock MIT Press, Cambridge, MA.

\bibitem[\protect\citename{Flekova and Gurevych}2016]{flekova:2016}
Flekova, L. and Gurevych, I.
\newblock (2016).
\newblock {Supersense Embeddings: A Unified Model for Supersense
  Interpretation, Prediction and Utilization}.
\newblock In {\em Proc. of ACL}, pages 2029--2041.

\bibitem[\protect\citename{Gale \bgroup et al.\egroup }1992]{galeetal:92}
Gale, W.~A., Church, K.~W., and Yarowsky, D.
\newblock (1992).
\newblock One sense per discourse.
\newblock In {\em Proceedings of the workshop on Speech and Natural Language},
  pages 233--237. Association for Computational Linguistics.

\bibitem[\protect\citename{Gonz{\'a}lez \bgroup et al.\egroup
  }2012]{gonzalez2012graph}
Gonz{\'a}lez, A., Rigau, G., and Castillo, M.
\newblock (2012).
\newblock A graph-based method to improve {W}ordnet domains.
\newblock In {\em Proceedings of 13th International Conference on Intelligent
  Text Processing and Computational Linguistics (CICLING)}, pages 17--28, New
  Delhi, India.

\bibitem[\protect\citename{Henrich \bgroup et al.\egroup
  }2012]{henrich2012webcage}
Henrich, V., Hinrichs, E., and Vodolazova, T.
\newblock (2012).
\newblock Webcage: a web-harvested corpus annotated with germanet senses.
\newblock In {\em Proceedings of the 13th Conference of the European Chapter of
  the Association for Computational Linguistics}, pages 387--396. Association
  for Computational Linguistics.

\bibitem[\protect\citename{Huang \bgroup et al.\egroup
  }2019]{huang2019glossbert}
Huang, L., Sun, C., Qiu, X., and Huang, X.-J.
\newblock (2019).
\newblock Glossbert: Bert for word sense disambiguation with gloss knowledge.
\newblock In {\em Proceedings of the 2019 Conference on Empirical Methods in
  Natural Language Processing and the 9th International Joint Conference on
  Natural Language Processing (EMNLP-IJCNLP)}, pages 3500--3505.

\bibitem[\protect\citename{Iacobacci \bgroup et al.\egroup
  }2015]{iacobacci:2015}
Iacobacci, I., Pilehvar, M.~T., and Navigli, R.
\newblock (2015).
\newblock {SensEmbed: Learning Sense Embeddings for Word and Relational
  Similarity}.
\newblock In {\em Proc. of ACL}, pages 95--105.

\bibitem[\protect\citename{Iacobacci \bgroup et al.\egroup
  }2016]{iacobaccietal:16}
Iacobacci, I., Pilehvar, M.~T., and Navigli, R.
\newblock (2016).
\newblock Embeddings for word sense disambiguation: An evaluation study.
\newblock In {\em Proceedings of the 54th Annual Meeting of the Association for
  Computational Linguistics}, pages 897--907.

\bibitem[\protect\citename{Ide \bgroup et al.\egroup }2002]{ideetal:02}
Ide, N., Erjavec, T., and Tufi{\c{s}}, D.
\newblock (2002).
\newblock Sense discrimination with parallel corpora.
\newblock In {\em Proceedings of the ACL-02 Workshop on {WSD}: Recent Successes
  and Future Directions, Philadelphia, Penn., July 2002}, pages 54--60.

\bibitem[\protect\citename{Ide \bgroup et al.\egroup }2008]{ideetal:08}
Ide, N., Baker, C., Fellbaum, C., Fillmore, C., and Passonneau, R.
\newblock (2008).
\newblock Masc: The manually annotated sub-corpus of american english.
\newblock In {\em 6th International Conference on Language Resources and
  Evaluation, LREC 2008}, pages 2455--2460. European Language Resources
  Association (ELRA).

\bibitem[\protect\citename{Ide \bgroup et al.\egroup
  }2010]{ide-etal-2010-manually}
Ide, N., Baker, C., Fellbaum, C., and Passonneau, R.
\newblock (2010).
\newblock The manually annotated sub-corpus: A community resource for and by
  the people.
\newblock In {\em Proceedings of the {ACL} 2010 Conference Short Papers}, pages
  68--73, Uppsala, Sweden, July. Association for Computational Linguistics.

\bibitem[\protect\citename{K{\aa}geb{\"{a}}ck and
  Salomonsson}2016]{kagelback:2016}
K{\aa}geb{\"{a}}ck, M. and Salomonsson, H.
\newblock (2016).
\newblock {Word Sense Disambiguation using a Bidirectional LSTM}.
\newblock In {\em Proc. of CogALex}, pages 51--56.

\bibitem[\protect\citename{Koehn}2005]{koehn05}
Koehn, P.
\newblock (2005).
\newblock Europarl: A parallel corpus for statistical machine translation.
\newblock In {\em Proceedings of Machine Translation Summit X}.

\bibitem[\protect\citename{Kucera and Francis}1979]{kucera1979standard}
Kucera, H. and Francis, W.
\newblock (1979).
\newblock A standard corpus of present-day edited american english, for use
  with digital computers (revised and amplified from 1967 version).

\bibitem[\protect\citename{Loureiro and
  Jorge}2019]{loureiro-jorge-2019-language}
Loureiro, D. and Jorge, A.
\newblock (2019).
\newblock Language modelling makes sense: Propagating representations through
  {W}ord{N}et for full-coverage word sense disambiguation.
\newblock In {\em Proceedings of the 57th Annual Meeting of the Association for
  Computational Linguistics}, pages 5682--5691, Florence, Italy, July.
  Association for Computational Linguistics.

\bibitem[\protect\citename{Luo \bgroup et al.\egroup
  }2018]{luo2018incorporating}
Luo, F., Liu, T., Xia, Q., Chang, B., and Sui, Z.
\newblock (2018).
\newblock Incorporating glosses into neural word sense disambiguation.
\newblock In {\em Proceedings of the 56th Annual Meeting of the Association for
  Computational Linguistics (Volume 1: Long Papers)}, volume~1, pages
  2473--2482.

\bibitem[\protect\citename{Mancini \bgroup et al.\egroup
  }2017]{mancini2017sw2v}
Mancini, M., Camacho{-}Collados, J., Iacobacci, I., and Navigli, R.
\newblock (2017).
\newblock Embedding words and senses together via joint knowledge-enhanced
  training.
\newblock In {\em Proceedings of CoNLL}, pages 100--111, Vancouver, Canada.

\bibitem[\protect\citename{Melamud \bgroup et al.\egroup }2016]{melamudetal:16}
Melamud, O., Goldberger, J., and Dagan, I.
\newblock (2016).
\newblock context2vec: Learning generic context embedding with bidirectional
  lstm.
\newblock In {\em Proceedings of CONLL}, pages 51--61.

\bibitem[\protect\citename{Miller \bgroup et al.\egroup }1993a]{Milleretal:93}
Miller, G.~A., Leacock, C., Tengi, R., and Bunker, R.
\newblock (1993a).
\newblock A semantic concordance.
\newblock In {\em Proceedings of the 3rd DARPA Workshop on Human Language
  Technology}, pages 303--308, Plainsboro, N.J.

\bibitem[\protect\citename{Miller \bgroup et al.\egroup }1993b]{miller93}
Miller, G.~A., Leacock, C., Tengi, R., and Bunker, R.
\newblock (1993b).
\newblock A semantic concordance.
\newblock In {\em Proceedings of the 3rd DARPA Workshop on Human Language
  Technology}, pages 303--308, Plainsboro, N.J.

\bibitem[\protect\citename{Moro and Navigli}2015]{MoroNavigli:15}
Moro, A. and Navigli, R.
\newblock (2015).
\newblock Semeval-2015 task 13: Multilingual all-words sense disambiguation and
  entity linking.
\newblock In {\em Proc. of SemEval-2015}.

\bibitem[\protect\citename{Moro \bgroup et al.\egroup }2014]{Moroetal:14tacl}
Moro, A., Raganato, A., and Navigli, R.
\newblock (2014).
\newblock {Entity Linking meets Word Sense Disambiguation: a Unified Approach}.
\newblock {\em Transaction of ACL (TACL)}, 2:231--244.

\bibitem[\protect\citename{Navigli and Ponzetto}2012]{NavigliPonzetto:12aij}
Navigli, R. and Ponzetto, S.~P.
\newblock (2012).
\newblock {B}abel{N}et: {T}he automatic construction, evaluation and
  application of a wide-coverage multilingual semantic network.
\newblock {\em Artificial Intelligence}, 193:217--250.

\bibitem[\protect\citename{Navigli \bgroup et al.\egroup }2013]{Naviglietal:13}
Navigli, R., Jurgens, D., and Vannella, D.
\newblock (2013).
\newblock Semeval-2013 task 12: Multilingual word sense disambiguation.
\newblock In {\em Proc. of SemEval 2013}, pages 222--231, Atlanta, USA.

\bibitem[\protect\citename{Navigli}2009]{navigli:09}
Navigli, R.
\newblock (2009).
\newblock {W}ord {S}ense {D}isambiguation: {A} survey.
\newblock {\em ACM Computing Surveys}, 41(2):1--69.

\bibitem[\protect\citename{Otegi \bgroup et al.\egroup }2016]{otegi:2016}
Otegi, A., Aranberri, N., Branco, A., Hajic, J., Neale, S., Osenova, P.,
  Pereira, R., Popel, M., Silva, J., Simov, K., and Agirre, E.
\newblock (2016).
\newblock {QTLeap WSD/NED Corpora: Semantic Annotation of Parallel Corpora in
  Six Languages}.
\newblock In {\em Proc. of LREC}, pages 3023--3030.

\bibitem[\protect\citename{Pasini and Navigli}2017]{pasininavigli:17}
Pasini, T. and Navigli, R.
\newblock (2017).
\newblock Train-o-matic: Large-scale supervised word sense disambiguation
  inmultiple languages without manual training data.
\newblock In {\em Proceedings of Empirical Methods in Natural Language
  Processing}, Copenhagen, Denmark.

\bibitem[\protect\citename{Pasini and Navigli}2018]{pasiniandnavigli:18}
Pasini, T. and Navigli, R.
\newblock (2018).
\newblock Two knowledge-based methods for high-performance sense distribution
  learning.
\newblock {\em Proceedings of AAAI, New Orleans, United States}.

\bibitem[\protect\citename{Pasini \bgroup et al.\egroup }2018]{pasinietal:18}
Pasini, T., Elia, F.~M., and Navigli, R.
\newblock (2018).
\newblock {Huge Automatically Extracted Training Sets for Multilingual Word
  Sense Disambiguation}.
\newblock In {\em Proceedings of LREC}, Miyazaki, Japan.

\bibitem[\protect\citename{Petrolito and Bond}2014]{petrolito2014survey}
Petrolito, T. and Bond, F.
\newblock (2014).
\newblock A survey of wordnet annotated corpora.
\newblock In {\em Proceedings Global WordNet Conference, GWC-2014}, pages
  236--245.

\bibitem[\protect\citename{Pilehvar \bgroup et al.\egroup
  }2013]{pilehvaretal:13}
Pilehvar, M.~T., Jurgens, D., and Navigli, R.
\newblock (2013).
\newblock Align, disambiguate and walk: A unified approach for measuring
  semantic similarity.
\newblock In {\em Proc. of ACL}, pages 1341--1351.

\bibitem[\protect\citename{Pradhan \bgroup et al.\egroup }2007]{Pradhanetal:07}
Pradhan, S.~S., Loper, E., Dligach, D., and Palmer, M.
\newblock (2007).
\newblock Semeval-2007 task 17: English lexical sample, srl and all words.
\newblock In {\em Proceedings of the 4th International Workshop on Semantic
  Evaluations}, pages 87--92.

\bibitem[\protect\citename{Raganato \bgroup et al.\egroup
  }2016]{raganatoetal:16}
Raganato, A., {Delli Bovi}, C., and Navigli, R.
\newblock (2016).
\newblock {Automatic Construction and Evaluation of a Large Semantically
  Enriched Wikipedia}.
\newblock In {\em Proceedings of IJCAI}, pages 2894--2900, New York City, NY,
  USA, July.

\bibitem[\protect\citename{Raganato \bgroup et al.\egroup
  }2017a]{raganatoetal:17}
Raganato, A., Camacho-Collados, J., and Navigli, R.
\newblock (2017a).
\newblock {Word Sense Disambiguation: A Unified Evaluation Framework and
  Empirical Comparison}.
\newblock In {\em Proc. of EACL}, pages 99--110, Valencia, Spain.

\bibitem[\protect\citename{Raganato \bgroup et al.\egroup
  }2017b]{raganatoemnlp2017}
Raganato, A., Delli~Bovi, C., and Navigli, R.
\newblock (2017b).
\newblock Neural sequence learning models for word sense disambiguation.
\newblock In {\em Proceedings of the 2017 Conference on Empirical Methods in
  Natural Language Processing}, pages 1167--1178. Association for Computational
  Linguistics.

\bibitem[\protect\citename{Scarlini \bgroup et al.\egroup
  }2019]{scarlinietal:19}
Scarlini, B., Pasini, T., and Navigli, R.
\newblock (2019).
\newblock {Just ''OneSeC`` for Producing Multilingual Sense-Annotated Data}.
\newblock In {\em Proceedings of the 57th Annual Meeting of the Association for
  Computational Linguistics}, pages 699--709.

\bibitem[\protect\citename{Schubert}2006]{schubert:2006}
Schubert, L.
\newblock (2006).
\newblock {Turing's Dream and the Knowledge Challenge}.
\newblock In {\em Proc. of AAAI}, pages 1534--1538.

\bibitem[\protect\citename{Singh \bgroup et al.\egroup
  }2012]{singh2012wikilinks}
Singh, S., Subramanya, A., Pereira, F., and McCallum, A.
\newblock (2012).
\newblock Wikilinks: A large-scale cross-document coreference corpus labeled
  via links to wikipedia.
\newblock {\em University of Massachusetts, Amherst, Technical Report
  UM-CS-2012- 015}.

\bibitem[\protect\citename{Snyder and Palmer}2004]{SnyderPalmer:04}
Snyder, B. and Palmer, M.
\newblock (2004).
\newblock The english all-words task.
\newblock In {\em Proc. of Senseval-3}, pages 41--43, Barcelona, Spain.

\bibitem[\protect\citename{Taghipour and Ng}2015a]{taghipourng:15}
Taghipour, K. and Ng, H.~T.
\newblock (2015a).
\newblock One million sense-tagged instances for word sense disambiguation and
  induction.
\newblock {\em CoNLL 2015}, pages 338--344.

\bibitem[\protect\citename{Taghipour and Ng}2015b]{taghipour2015semi}
Taghipour, K. and Ng, H.~T.
\newblock (2015b).
\newblock {Semi-Supervised Word Sense Disambiguation Using Word Embeddings in
  General and Specific Domains}.
\newblock {\em Proc. of NAACL-HLT}, pages 314--323.

\bibitem[\protect\citename{Usbeck \bgroup et al.\egroup
  }2015]{usbeck2015gerbil}
Usbeck, R., R{\"o}der, M., Ngonga~Ngomo, A.-C., Baron, C., Both, A.,
  Br{\"u}mmer, M., Ceccarelli, D., Cornolti, M., Cherix, D., Eickmann, B.,
  et~al.
\newblock (2015).
\newblock Gerbil: general entity annotator benchmarking framework.
\newblock In {\em Proceedings of the 24th International Conference on World
  Wide Web}, pages 1133--1143. International World Wide Web Conferences
  Steering Committee.

\bibitem[\protect\citename{Vial \bgroup et al.\egroup }2018]{vialetal:18}
Vial, L., Lecouteux, B., and Schwab, D.
\newblock (2018).
\newblock {UFSAC: Unification of Sense Annotated Corpora and Tools}.
\newblock In Nicoletta Calzolari~(Conference chair), et~al., editors, {\em
  Proceedings of the Eleventh International Conference on Language Resources
  and Evaluation (LREC 2018)}, Miyazaki, Japan, May 7-12, 2018. European
  Language Resources Association (ELRA).

\bibitem[\protect\citename{Weischedel \bgroup et al.\egroup
  }2013]{weischedel2013ontonotes}
Weischedel, R., Palmer, M., Marcus, M., Hovy, E., Pradhan, S., Ramshaw, L.,
  Xue, N., Taylor, A., Kaufman, J., Franchini, M., et~al.
\newblock (2013).
\newblock Ontonotes release 5.0.
\newblock {\em Linguistic Data Consortium, Philadelphia, PA}.

\bibitem[\protect\citename{Yuan \bgroup et al.\egroup }2016]{yuan:16}
Yuan, D., Richardson, J., Doherty, R., Evans, C., and Altendorf, E.
\newblock (2016).
\newblock Semi-supervised word sense disambiguation with neural models.
\newblock {\em Proceedings of COLING}, pages 1374--1385.

\bibitem[\protect\citename{Zhong and Ng}2010]{zhongng:10}
Zhong, Z. and Ng, H.~T.
\newblock (2010).
\newblock It makes sense: A wide-coverage word sense disambiguation system for
  free text.
\newblock In {\em Proc. of of the ACL}, pages 78--83, Uppsala, Sweden. ACL.

\end{thebibliography}


\end{document}